\documentclass[11pt]{article}

\usepackage[preprint]{acl}

\usepackage{times}
\usepackage{latexsym}

\usepackage{verbatim}
\usepackage{tcolorbox}
\tcbuselibrary{skins,breakable}
\usepackage{fvextra}
\usepackage{url}
\usepackage{hyperref}
\usepackage{booktabs}
\usepackage{amsmath}
\usepackage{enumitem}
\usepackage{amssymb}
\usepackage[table]{xcolor}
\usepackage{subcaption}
\usepackage{float}
\usepackage[T1]{fontenc}

\usepackage[utf8]{inputenc}

\usepackage{microtype}

\usepackage{inconsolata}

\usepackage{graphicx}

\title{Learning When to Reason for Text-to-SQL via SFT and DPO}

\author{
  \textbf{Soohyuk Jang\textsuperscript{1}},
  \textbf{Jiheum Yeom\textsuperscript{1}},
  \textbf{Nohil Park\textsuperscript{1}},
\\
  \textbf{Sang Hun Kim\textsuperscript{2}},
  \textbf{Yoonyoung Choi\textsuperscript{2}},
  \textbf{Kiwook Bae\textsuperscript{2}},
  \textbf{Sungroh Yoon\textsuperscript{1,3}}\thanks{Corresponding author.}
\\
\\
  \textsuperscript{1}Department of Electrical and Computer Engineering, Seoul National University \\
  \textsuperscript{2}AI Center, Samsung Electronics \\
  \textsuperscript{3}AIIS, ASRI, INMC, ISRC, and IPAI, Seoul National University
\\
  {\texttt{\{soohyuk.jang, quilava1234, pnoil2588, sryoon\}@snu.ac.kr}} \\
  {\texttt{\{phd.kim, yy45.choi, kiwook.bae\}@samsung.com}}
}

\begin{document}
\maketitle

\begin{abstract}
Recent Text-to-SQL methods rely heavily on reasoning-centric paradigms such as Chain-of-Thought (CoT), achieving substantial gains on complex benchmarks at the cost of high inference-time overhead. 
However, a large fraction of real-world queries are simple lookups or aggregations that can be resolved without multi-step deduction, making forced reasoning wasteful.
Thus, we propose \textbf{AutoThinkSQL}, a framework that integrates an auto-thinking mechanism into both Supervised Fine-Tuning (SFT) and Direct Preference Optimization (DPO) on Text-to-SQL. 
Our approach enables the model to dynamically bypass reasoning for simple queries while invoking deep CoT for complex queries.
On Qwen3-Coder-30B-A3B, our method achieves consistent gains compared to the best counterpart baseline on both Spider and BIRD benchmarks while simultaneously reducing average output tokens by 24.6\% and 18.3\%, and average latency by 17.1\% and 11.5\% compared to CoT-only generation.
Further analysis indicates that the model learns to align its reasoning decisions with query difficulty.
\end{abstract}

\section{Introduction}\label{sec1:introduction}

Text-to-SQL translates natural language questions into executable SQL queries, providing an intuitive interface for relational databases~\citep{omnisql, codes, dinsql, uncoveringdpo, qwen25coder}.
While Large Language Models (LLMs) have driven significant advancements in this field, recent state-of-the-art methods heavily rely on reasoning-centric learning paradigms, driven by Reinforcement Learning (RL)~\citep{arctic, sqlr1, reasoningsql} and Chain-of-Thought (CoT) prompting~\citep{dinsql}. 
By explicitly generating intermediate logical steps, these approaches effectively bridge the semantic gap between complex user intents and intricate database schemas, yielding substantial accuracy gains on challenging benchmarks~\citep{reasoningsql, arctic, sqlr1}.

However, this reliance on verbose reasoning traces introduces a critical inference-time bottleneck. 
This is particularly problematic given that a substantial fraction of real-world Text-to-SQL queries are simple lookups or single-table aggregations that can be resolved through straightforward schema linking~\citep{spider, bird}. 
Compelling a model to produce extensive reasoning traces for such queries not only wastes inference budget but can also introduce syntactic noise and hallucinations~\citep{uncoveringdpo, cotnotcot}.
This motivates a training strategy that retains CoT's accuracy benefits on complex queries while avoiding its inference cost on simpler ones.

To address this, we introduce \textbf{AutoThinkSQL}, a novel framework that integrates an adaptive auto-thinking~\citep{lou2025adacot,tu2026learning} mechanism into the Text-to-SQL domain. 
Our approach equips LLM with the intrinsic ability to bypass CoT for simple queries and reserve it strictly for complex reasoning tasks. 
We implement this through a two-stage training pipeline of SFT and DPO~\citep{dpo}. 
Evaluated on Qwen3-Coder-30B-A3B~\citep{qwen3coder}, our auto-thinking SFT consistently surpasses the execution accuracy of its CoT-only counterpart across most settings, while significantly reducing the average number of inference tokens and latency.

Our primary contributions are threefold:
\begin{itemize}[itemsep=0.1pt]
    \item \textbf{Consistent gains.}
    AutoThinkSQL achieves strong performance across both Spider and BIRD under both decoding regimes, while single-mode baselines each show a clear weakness in at least one setting.
    \item \textbf{Inference efficiency.}
    AutoThinkSQL reduces average output tokens by 24.6\% on Spider and 18.3\% on BIRD, and average latency by 17.1\% on Spider and 11.5\% on BIRD, compared to the CoT-only (SFT+DPO) counterpart, substantially lowering inference-time cost without sacrificing accuracy.
    \item \textbf{Routing analysis.}
    We analyze the model's reasoning activation patterns and verify that AutoThinkSQL adaptively routes queries according to difficulty, confirming that the efficiency gains reflect meaningful behavioral alignment with query complexity.
\end{itemize}
\section{Related Work}\label{sec:related_work}

Prior work on LLM-based Text-to-SQL falls into two broad categories. 
One line leverages proprietary LLMs without parameter updates through prompt engineering and inference-time reasoning, including schema-linking decomposition~\citep{dinsql}, in-context example selection~\citep{gao2024dailsql}, multi-agent collaboration~\citep{wang-etal-2025-mac, talaei2024chess}, and multi-path candidate selection~\citep{pourreza2025chasesql}. 
The other fine-tunes open-source LLMs to close the gap with proprietary models, ranging from incremental pre-training~\citep{codes} and large-scale synthetic SFT~\citep{omnisql} to DPO-based alignment~\citep{yang-etal-2024-synthesizing} and RL with reasoning traces~\citep{reasoningsql}. 
Despite their effectiveness, these fine-tuning approaches force chain-of-thought reasoning regardless of query difficulty, inflating the inference-time token budget even on queries that could be resolved without multi-step deduction.

A parallel line of work on adaptive reasoning trains models to decide when to invoke CoT, using either SFT followed by PPO~\citep{lou2025adacot, ppo} or GRPO-based RL~\citep{zhang-etal-2025-adaptthink,
fang2026thinkless, tu2026learning}. 
However, these methods usually target math and general reasoning, and to our knowledge no prior work has trained Text-to-SQL models to adaptively decide when to reason. 
As a closest related attempt, \citet{tai-etal-2023-exploring} observed that detailed reasoning can amplify errors on simpler SQL queries, but provided only a prompting-level fix without addressing it during training. 

\section{Methodology}\label{sec2:method}

\subsection{Problem Formulation}\label{ssec21:formulation}
Given a natural language question, a database schema, and external knowledge, the Text-to-SQL task aims to generate a valid SQL query.

We denote the input comprising all three as $x$ (Figure~\ref{fig:shared_template}), and introduce mode-specific format prompts $p_m$ that govern the output structure.
Specifically, $p_{\text{NC}}$ instructs the model to generate SQL directly without reasoning (Figure~\ref{fig:format_nocot}), $p_{\text{C}}$ requires step-by-step reasoning prior to the final SQL (Figure~\ref{fig:format_cot}), and $p_{\text{Auto}}$ instructs the model to first assess query complexity and then either generate SQL directly or produce a reasoning chain before the SQL (Figure~\ref{fig:format_ours}).
Given $x$ and a prompt $p_m$, the model produces a response $y_m$, i.e., $(x, p_m) \rightarrow y_m$.

\subsection{Auto-thinking SFT}\label{ssec22:sft}

\paragraph{Data Construction.}
For each BIRD training instance $x$, we execute $16$ independent rollouts under each mode-specific format prompt:
$(x, p_{\text{NC}})$ yields $\{y_{\text{NC}}^{(i)}\}_{i=1}^{16}$ and $(x, p_{\text{C}})$ yields $\{y_{\text{C}}^{(i)}\}_{i=1}^{16}$.
Let $N_{\text{C}}$ and $N_{\text{NC}}$ denote the number of correct outputs under each mode.
The training mode $m^*$ is assigned as follows:
\[
m^* = \begin{cases}
\text{NC}       & \text{if } N_{\text{NC}} = 16 \\
\text{C}        & \text{if } N_{\text{NC}} < 16 \wedge N_{\text{C}} > 0 \\
\text{Discard} & \text{otherwise}
\end{cases}
\]
For each labeled instance, one correct rollout is uniformly sampled from mode $m^*$, yielding $y_{m^*}$.
The pair $((x,\, p_{\text{Auto}}),\, y_{m^*})$ is then added to $\mathcal{D}_{\text{SFT}}$.
Although rollouts are collected under $p_{\text{NC}}$ and $p_{\text{C}}$, the SFT input is always composed with $p_{\text{Auto}}$, so the model learns to autonomously produce the appropriate output style.

\paragraph{Training Objective.}
The model is fine-tuned by minimizing the standard negative log-likelihood over $\mathcal{D}_{\text{SFT}}$:
\begin{equation*}
\begin{aligned}
\mathcal{L}_{\text{SFT}}(\theta) = & -\mathbb{E}_{((x, p_{\text{Auto}}), y) \sim \mathcal{D}_{\text{SFT}}} \\
  & \sum_{t=1}^{|y|} \log p_\theta(y_t \mid x, p_{\text{Auto}}, y_{<t})
\end{aligned}
\end{equation*}
where $y$ is the target output trajectory, which may include a reasoning chain followed by the SQL query.

\subsection{Auto-thinking DPO}\label{ssec23:dpo}
\paragraph{Data Construction.}
Using the SFT-trained model, we collect $16$ new rollouts per mode under the same prompts $(p_{\text{NC}},\, p_{\text{C}})$ as in Section~\ref{ssec22:sft}.
Let $y_m^+$ and $y_m^-$ denote a uniformly sampled chosen and rejected response from mode $m \in \{\text{C}, \text{NC}\}$, respectively.
Preference pairs $(y_w, y_l)$ are assigned according to the following mutually exclusive, exhaustive rule:
\[
\resizebox{\columnwidth}{!}{$\displaystyle
(y_w,\, y_l) = \begin{cases}
\text{Discard}
  & \text{if } N_{\text{C}} = 0 \wedge N_{\text{NC}} = 0 \\[2pt]
\bigl(y_{\text{NC}}^+,\; y_{\text{C}}^-\bigr)
  & \text{if } N_{\text{NC}} = 16 \wedge N_{\text{C}} < 16 \\[2pt]
\bigl(y_{\text{NC}}^+,\; y_{\text{C}}^+\bigr)
  & \text{if } N_{\text{NC}} = 16 \wedge N_{\text{C}} = 16 \\[2pt]
\begin{aligned}[t] 
  & \bigl(y_{m}^+,\; y_{m}^-\bigr), \\ 
  & \quad m \sim \mathrm{Unif}(\{\text{C},\text{NC}\}) 
\end{aligned}
  & \text{if } 0 < N_{\text{C}} = N_{\text{NC}} < 16 \\[2pt]
\bigl(y_{\text{NC}}^+,\; y_{\text{C}}^-\bigr)
  & \text{if } N_{\text{C}} < N_{\text{NC}} < 16 \\[2pt]
\bigl(y_{\text{C}}^+,\; y_{\text{NC}}^-\bigr)
  & \text{if } N_{\text{NC}} < N_{\text{C}}
\end{cases}
$}
\]

When $N_{\text{NC}} = 16$, the No-CoT response is set as $y_w$ to discourage unnecessary reasoning on trivially solvable queries.
Each accepted preference pair $((x,\, p_{\text{Auto}}),\, y_w,\, y_l)$ is added to $\mathcal{D}_{\text{DPO}}$.

\paragraph{Training Objective.}
The model is further optimized using the standard DPO objective over $\mathcal{D}_{\text{DPO}}$:
\begin{equation*}
\begin{aligned}
\mathcal{L}_{\text{DPO}}(\theta) = -\mathbb{E}_{\mathcal{D}_{\text{DPO}}} \log \sigma\Big( 
  & \beta \log \tfrac{p_\theta(y_w \mid x, p_{\text{Auto}})}{p_{\text{ref}}(y_w \mid x, p_{\text{Auto}})} \\
  & - \beta \log \tfrac{p_\theta(y_l \mid x, p_{\text{Auto}})}{p_{\text{ref}}(y_l \mid x, p_{\text{Auto}})}
  \Big)
\end{aligned}
\end{equation*}
where $p_{\text{ref}}$ is the frozen SFT reference policy and $\beta$ controls deviation from the reference.

\section{Experiments}\label{sec3:experiments}

\subsection{Experimental Setup}\label{ssec31:setup}

\paragraph{Models.}
We conduct experiments on Qwen3-Coder-30B-A3B-Instruct\footnote{\href{https://huggingface.co/Qwen/Qwen3-Coder-30B-A3B-Instruct}{huggingface.co/Qwen/Qwen3-Coder-30B-A3B-Instruct}}~\cite{qwen3coder}, a Mixture-of-Experts model with 30B total and 3B activated parameters, as our backbone.
All fine-tuning is performed with LoRA~\citep{LoRA}; full training details are in Appendix~\ref{app:impl}.

\paragraph{Training Data.}
We construct our SFT and DPO datasets from the BIRD~\citep{bird} training set which consists 9428 samples following the mode labeling procedure described in Sections~\ref{ssec22:sft} and \ref{ssec23:dpo}.
For the database prompt construction, we adopt the preprocessed schema format from \citet{uncoveringdpo}, which follows the CodeS~\citep{codes} pipeline.
Detailed information on the schema preprocessing procedure and dataset statistics are provided in Appendix~\ref{app:preprocessing} (Table~\ref{tab:sft_data_stats}, Table~\ref{tab:dpo_data_stats}).

\paragraph{Evaluation Benchmarks.}
We evaluate on Spider~\citep{spider}, a cross-domain benchmark with queries spanning four difficulty levels, and 
BIRD~\citep{bird}, which targets more realistic queries requiring external knowledge and exhibits heavier difficulty skew. 

\paragraph{Evaluation Metric.}
We report \textit{Execution Accuracy} (EX), the standard Text-to-SQL metric measuring the percentage of queries whose execution results match the gold SQL.

\paragraph{Decoding Strategy.}
We evaluate each model under two decoding settings.
\textit{Greedy decoding} selects the most probable token at each step.
\textit{Majority voting at 8} (Maj@8) generates 8 independent outputs using sampling (temperature $=1.0$) and selects the final answer by majority vote over the resulting SQL execution outcomes.

\begin{table}[t]
    \centering
    \caption{Performance comparison of different models on the Spider and BIRD development sets. Both Greedy and Maj@8 decoding results are reported (mean$\pm$std over 3 runs). The best and second-best results per column are shown in \textbf{bold} and \underline{underlined}, respectively.}
    \label{tab:main_results}
    \resizebox{\columnwidth}{!}{%
    \begin{tabular}{l c c c c}
        \toprule
        \textbf{Model} & \multicolumn{2}{c}{\textbf{Spider Dev}} & \multicolumn{2}{c}{\textbf{BIRD Dev}} \\
        \cmidrule(lr){2-3} \cmidrule(lr){4-5}
                   & \textbf{Greedy} & \textbf{Maj@8} & \textbf{Greedy} & \textbf{Maj@8} \\
        \midrule
        Qwen3-Coder-30B-A3B & $82.49_{\pm0.19}$ & $82.81_{\pm0.31}$ & $55.84_{\pm0.23}$ & $56.21_{\pm0.23}$ \\
        \midrule
        \rowcolor{gray!15} \multicolumn{5}{l}{\textit{SFT}} \\
        \rowcolor{gray!15} ~~No-CoT             & $83.81_{\pm0.15}$ & $84.20_{\pm0.15}$ & $56.91_{\pm0.46}$ & $57.79_{\pm0.27}$ \\
        \rowcolor{gray!15} ~~CoT                & $84.59_{\pm0.58}$ & $\mathbf{87.07_{\pm0.05}}$ & $56.88_{\pm0.44}$ & $59.80_{\pm0.56}$ \\
        \rowcolor{gray!15} ~~\textbf{AutoThinkSQL}      & $\underline{85.26_{\pm0.20}}$ & $86.68_{\pm0.30}$ & $57.15_{\pm0.30}$ & $\underline{60.23_{\pm0.58}}$ \\
        \midrule
        \rowcolor{blue!5} \multicolumn{5}{l}{\textit{SFT + DPO}} \\
        \rowcolor{blue!5} ~~No-CoT       & $84.23_{\pm0.10}$ & $84.46_{\pm0.20}$ & $\underline{57.64_{\pm0.10}}$ & $58.01_{\pm0.45}$ \\
        \rowcolor{blue!5} ~~CoT          & $69.14_{\pm0.54}$ & $83.33_{\pm0.58}$ & $54.02_{\pm0.76}$ & $59.55_{\pm0.64}$ \\
        \rowcolor{blue!5} ~~\textbf{AutoThinkSQL} & $\mathbf{86.20_{\pm0.25}}$ & $\underline{86.78_{\pm0.11}}$ & $\mathbf{58.23_{\pm0.61}}$ & $\mathbf{60.27_{\pm0.30}}$ \\
        \bottomrule
    \end{tabular}%
    }
\end{table}

\paragraph{Baselines.}
We compare AutoThinkSQL SFT and AutoThinkSQL SFT+DPO (ours) against five baselines: 
\textit{Zero-shot}, the base Qwen3-Coder-30B-A3B-Instruct~\cite{qwen3coder} model; 
and \textit{CoT-only} and \textit{No-CoT-only} variants under both SFT and SFT+DPO, which are trained to always generate or always skip a reasoning trace, respectively.
Full training hyperparameters are provided in Appendix~\ref{app:impl}.
The prompt templates used for all training and evaluation are provided in Appendix~\ref{sec:appendix_prompts}.

\subsection{Main Results}\label{ssec32:main}

Table~\ref{tab:main_results} reports Execution Accuracy on Spider~\cite{spider} and BIRD~\cite{bird} dev sets consisting of 1034 and 1534 samples respectively across all baselines and our proposed methods, averaged over three runs with standard deviations.

\paragraph{AutoThinkSQL performs strongly across most settings.}
AutoThinkSQL SFT+DPO records the highest execution accuracy in three of four evaluation settings: Spider Greedy (86.20), BIRD Greedy (58.23), and BIRD Maj@8 (60.27).
On Spider Maj@8, AutoThinkSQL SFT+DPO reaches 86.78, which is $0.29$ points behind CoT-only SFT (87.07), a gap within one standard deviation.
At the SFT stage alone, AutoThinkSQL already outperforms or matches both single-mode baselines, ranking first or second across all evaluated settings.

\paragraph{Single-mode training reveals decoding-dependent limitations.}
No-CoT-only SFT yields moderate gains under greedy decoding on both benchmarks (Spider: $83.81$, BIRD: $56.91$), yet its Maj@8 scores lag noticeably behind CoT-only SFT by $2.87$ points on Spider and $2.01$ on BIRD, as the lack of reasoning diversity limits the benefit of majority voting.
CoT-only SFT shows the opposite pattern, achieving the highest Maj@8 on Spider ($87.07$) but becoming less stable on greedy decoding once DPO is applied on top.
These contrasting behaviors suggest that committing to a single reasoning mode makes it difficult to remain competitive across both decoding regimes.
In contrast, AutoThinkSQL outperforms the best counterpart among baselines on both Spider and Bird in most cases, demonstrating consistent gains across both benchmarks.

\paragraph{AutoThinkSQL generalizes consistently across benchmarks.}
While single-mode baselines each exhibit a clear weakness under at least one evaluation setting, AutoThinkSQL improves consistently under both greedy and Maj@8 decoding, and on both Spider, which has a relatively balanced difficulty distribution, and BIRD, which contains heavier difficulty skew and requires external knowledge.

\begin{table}[t]
    \centering
    \small
    \caption{Inference efficiency comparison on Spider and BIRD development sets (mean$\pm$std over 3 runs).
    \textit{Lat.}\ denotes average latency (ms) per problem,
    and \textit{Tok.}\ denotes average output tokens per generation.}
    \label{tab:efficiency}
    \resizebox{\columnwidth}{!}{%
    \begin{tabular}{l rr rr}
        \toprule
        & \multicolumn{2}{c}{\textbf{Spider Dev}} & \multicolumn{2}{c}{\textbf{BIRD Dev}} \\
        \cmidrule(lr){2-3} \cmidrule(lr){4-5}
        \textbf{Model} & \multicolumn{1}{c}{\textit{Lat.\ (ms)}} & \multicolumn{1}{c}{\textit{Tok.}} & \multicolumn{1}{c}{\textit{Lat.\ (ms)}} & \multicolumn{1}{c}{\textit{Tok.}} \\
        \midrule
        Qwen3-Coder-30B-A3B & $72.0_{\pm0.1}$ & $32.7_{\pm0.1}$ & $119.6_{\pm0.6}$ & $52.1_{\pm0.1}$ \\
        \midrule
        \rowcolor{gray!15} \multicolumn{5}{l}{\textit{SFT}} \\
        \rowcolor{gray!15} ~~No-CoT           & $80.3_{\pm0.2}$  & $35.0_{\pm0.1}$ & $145.7_{\pm0.6}$  & $64.3_{\pm0.0}$ \\
        \rowcolor{gray!15} ~~CoT              & $454.7_{\pm2.3}$ & $236.4_{\pm0.3}$ & $604.0_{\pm3.3}$ & $320.5_{\pm1.0}$ \\
        \rowcolor{gray!15} ~~\textbf{AutoThinkSQL}    & $393.2_{\pm3.0}$ & $187.0_{\pm0.7}$ & $557.7_{\pm2.8}$ & $275.1_{\pm1.0}$ \\
        \midrule
        \rowcolor{blue!5} \multicolumn{5}{l}{\textit{SFT + DPO}} \\
        \rowcolor{blue!5} ~~No-CoT           & $73.7_{\pm3.5}$  & $32.5_{\pm0.0}$ & $127.3_{\pm0.8}$  & $54.8_{\pm0.4}$ \\
        \rowcolor{blue!5} ~~CoT        & $448.5_{\pm1.3}$ & $233.7_{\pm0.4}$ & $613.6_{\pm1.5}$ & $324.8_{\pm1.0}$ \\
        \rowcolor{blue!5} ~~\textbf{AutoThinkSQL} & $371.7_{\pm2.1}$ & $176.2_{\pm0.8}$ & $543.3_{\pm2.8}$ & $265.2_{\pm1.3}$ \\
        \bottomrule
    \end{tabular}%
    }
\end{table}

\subsection{Analysis}\label{ssec33:analysis}

\begin{figure}
    \centering
    \includegraphics[width=\columnwidth]{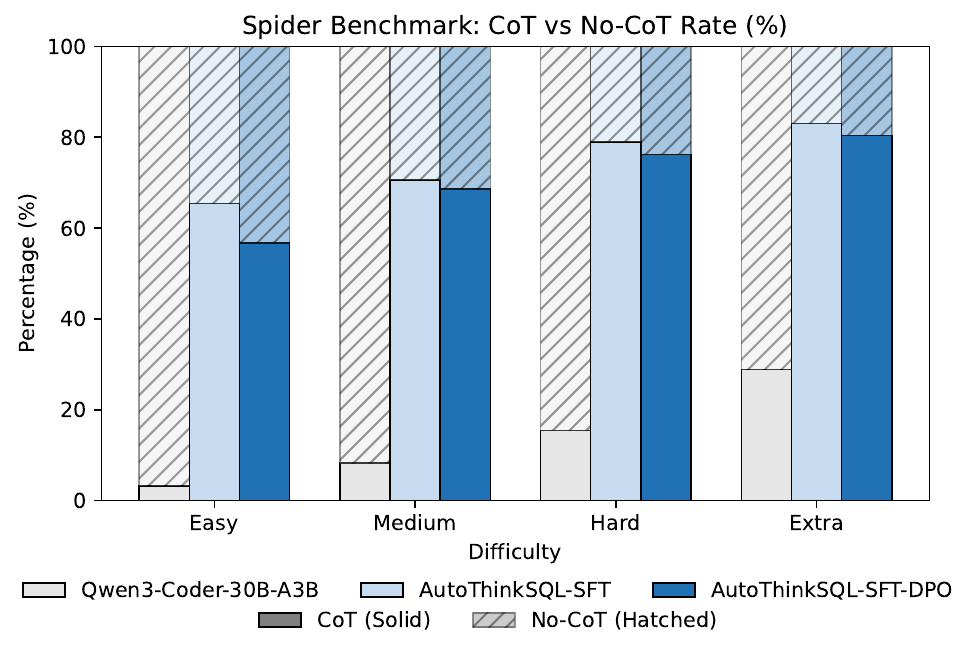}
    \caption{Comparison of Chain-of-Thought (CoT) and No-CoT trigger rates on the Spider~\cite{spider} benchmark across different difficulty levels. The rates were calculated by generating 8 independent responses for each question.}
    \label{fig:trigger_rate}
\end{figure}

We analyze the inference cost savings enabled by auto-thinking (Table~\ref{tab:efficiency}) and verify that routing behavior aligns with query difficulty (Figure~\ref{fig:trigger_rate}).

\paragraph{AutoThinkSQL substantially reduces inference cost.}
Compared to CoT-only SFT, AutoThinkSQL SFT reduces average output tokens by 21.0\% on Spider and 14.7\% on BIRD, while matching or exceeding its accuracy.
AutoThinkSQL SFT+DPO uses 24.9\% fewer tokens on Spider and 18.1\% fewer on BIRD than CoT-only SFT+DPO, with latency following the same trend.

\paragraph{Routing aligns with query difficulty.}
Figure~\ref{fig:trigger_rate} reports the CoT trigger rate of AutoThinkSQL across Spider's four difficulty levels (Easy, Medium, Hard, Extra Hard) as defined by \citet{spider}, computed over 8 independent samples per query.
Both AutoThinkSQL SFT and SFT+DPO show a consistent monotonic trend: as query difficulty increases, the No-CoT rate decreases and the CoT rate rises correspondingly, confirming that the model allocates reasoning effort in accordance with query complexity.
A zero-shot baseline applying our auto-thinking prompt (Figure~\ref{fig:format_ours}) without fine-tuning shows the same qualitative trend but at substantially lower CoT rates, indicating that the base model under-invokes reasoning.

\section{Conclusion}\label{sec4:conclusion}

We proposed AutoThinkSQL, the first Text-to-SQL framework to integrate auto-thinking into both SFT and DPO.
By learning when to invoke CoT, AutoThinkSQL matches or surpasses CoT-only baselines on Spider and BIRD while substantially reducing inference cost, and remains robust under preference-based alignment where single-mode training degrades.
We leave extension to larger models and more diverse Text-to-SQL data as future work.

\section*{Limitations}
In this work, our primary evaluation focuses on Spider and BIRD; we leave the exploration of other dialects, such as Spider 2.0~\citep{lei2025spider} or BIRD-CRITIC~\citep{li2026swesql}, as an avenue for future work. For mode assignment, we employ rollout-based supervision, which opens up opportunities for future research into more computationally lightweight heuristics at scale. Lastly, we concentrate our current experiments on the Qwen3-Coder family, paving the way for subsequent studies to examine cross-family generalization to other code-specialized LLMs like DeepSeek-Coder~\citep{guo2024deepseek}.

\section*{Potential Risks}
Like other Text-to-SQL systems, our model can generate incorrect SQL that, if executed without review, may cause unintended data access or modification in production databases. 
An additional risk specific to our framework is that the auto-thinking mechanism may bypass reasoning for queries it incorrectly deems simple, reducing the interpretability of the generated output and making errors harder to detect. 
We recommend human review or sandboxed execution in deployment, particularly for high-stakes use cases.

\section*{Acknowledgments}

This work was supported by the National Research Foundation of Korea (NRF) grant funded by the Korea government (MSIT) (No. 2022R1A3B1077720), the BK21 FOUR program of the Education and Research Program for Future ICT Pioneers, Seoul National University in 2026, Institute of Information \& communications Technology Planning \& Evaluation (IITP) grant funded by the Korea government(MSIT) [No.RS-2021-II211343, Artificial Intelligence Graduate School Program (Seoul National University), No.2022-0-00959, RS-2022-II220959], and Samsung Electronics Co., Ltd [IO250624-13143-01].

\bibliography{custom}

@inproceedings{dinsql,
    title={{DIN}-{SQL}: Decomposed In-Context Learning of Text-to-{SQL} with Self-Correction},
    author={Mohammadreza Pourreza and Davood Rafiei},
    booktitle={Thirty-seventh Conference on Neural Information Processing Systems},
    year={2023},
    url={https://openreview.net/forum?id=p53QDxSIc5}
}

@article{gao2024dailsql,
    author = {Gao, Dawei and Wang, Haibin and Li, Yaliang and Sun, Xiuyu and Qian, Yichen and Ding, Bolin and Zhou, Jingren},
    title = {Text-to-SQL Empowered by Large Language Models: A Benchmark Evaluation},
    year = {2024},
    issue_date = {January 2024},
    publisher = {VLDB Endowment},
    volume = {17},
    number = {5},
    issn = {2150-8097},
    url = {https://doi.org/10.14778/3641204.3641221},
    doi = {10.14778/3641204.3641221},
    journal = {Proc. VLDB Endow.},
    month = jan,
    pages = {1132–1145},
    numpages = {14}
}

@inproceedings{wang-etal-2025-mac,
    title = "{MAC}-{SQL}: A Multi-Agent Collaborative Framework for Text-to-{SQL}",
    author = "Wang, Bing  and
      Ren, Changyu  and
      Yang, Jian  and
      Liang, Xinnian  and
      Bai, Jiaqi  and
      Chai, LinZheng  and
      Yan, Zhao  and
      Zhang, Qian-Wen  and
      Yin, Di  and
      Sun, Xing  and
      Li, Zhoujun",
    editor = "Rambow, Owen  and
      Wanner, Leo  and
      Apidianaki, Marianna  and
      Al-Khalifa, Hend  and
      Eugenio, Barbara Di  and
      Schockaert, Steven",
    booktitle = "Proceedings of the 31st International Conference on Computational Linguistics",
    month = jan,
    year = "2025",
    address = "Abu Dhabi, UAE",
    publisher = "Association for Computational Linguistics",
    url = "https://aclanthology.org/2025.coling-main.36/",
    pages = "540--557"
}

@article{talaei2024chess,
    title={Chess: Contextual harnessing for efficient sql synthesis},
    author={Talaei, Shayan and Pourreza, Mohammadreza and Chang, Yu-Chen and Mirhoseini, Azalia and Saberi, Amin},
    journal={arXiv preprint arXiv:2405.16755},
    year={2024}
}

@inproceedings{pourreza2025chasesql,
    title={{CHASE}-{SQL}: Multi-Path Reasoning and Preference Optimized Candidate Selection in Text-to-{SQL}},
    author={Mohammadreza Pourreza and Hailong Li and Ruoxi Sun and Yeounoh Chung and Shayan Talaei and Gaurav Tarlok Kakkar and Yu Gan and Amin Saberi and Fatma Ozcan and Sercan O Arik},
    booktitle={The Thirteenth International Conference on Learning Representations},
    year={2025},
    url={https://openreview.net/forum?id=CvGqMD5OtX}
}

@inproceedings{yang-etal-2024-synthesizing,
    title = "Synthesizing Text-to-{SQL} Data from Weak and Strong {LLM}s",
    author = "Yang, Jiaxi  and
      Hui, Binyuan  and
      Yang, Min  and
      Yang, Jian  and
      Lin, Junyang  and
      Zhou, Chang",
    editor = "Ku, Lun-Wei  and
      Martins, Andre  and
      Srikumar, Vivek",
    booktitle = "Proceedings of the 62nd Annual Meeting of the Association for Computational Linguistics (Volume 1: Long Papers)",
    month = aug,
    year = "2024",
    address = "Bangkok, Thailand",
    publisher = "Association for Computational Linguistics",
    url = "https://aclanthology.org/2024.acl-long.425/",
    doi = "10.18653/v1/2024.acl-long.425",
    pages = "7864--7875"
}

@article{lou2025adacot,
    title={Adacot: Pareto-optimal adaptive chain-of-thought triggering via reinforcement learning},
    author={Lou, Chenwei and Sun, Zewei and Liang, Xinnian and Qu, Meng and Shen, Wei and Wang, Wenqi and Li, Yuntao and Yang, Qingping and Wu, Shuangzhi},
    journal={arXiv preprint arXiv:2505.11896},
    year={2025}
}

@inproceedings{zhang-etal-2025-adaptthink,
    title = "{A}dapt{T}hink: Reasoning Models Can Learn When to Think",
    author = "Zhang, Jiajie  and
      Lin, Nianyi  and
      Hou, Lei  and
      Feng, Ling  and
      Li, Juanzi",
    editor = "Christodoulopoulos, Christos  and
      Chakraborty, Tanmoy  and
      Rose, Carolyn  and
      Peng, Violet",
    booktitle = "Proceedings of the 2025 Conference on Empirical Methods in Natural Language Processing",
    month = nov,
    year = "2025",
    address = "Suzhou, China",
    publisher = "Association for Computational Linguistics",
    url = "https://aclanthology.org/2025.emnlp-main.184/",
    doi = "10.18653/v1/2025.emnlp-main.184",
    pages = "3716--3730",
    ISBN = "979-8-89176-332-6"
}

@inproceedings{fang2026thinkless,
    title={Thinkless: {LLM} Learns When to Think},
    author={Gongfan Fang and Xinyin Ma and Xinchao Wang},
    booktitle={The Thirty-ninth Annual Conference on Neural Information Processing Systems},
    year={2026},
    url={https://openreview.net/forum?id=ariVQf0KZx}
}

@inproceedings{tu2026learning,
    title={Learning When to Think: Shaping Adaptive Reasoning in R1-Style Models via Multi-Stage {RL}},
    author={Songjun Tu and Jiahao Lin and Qichao Zhang and Xiangyu Tian and Linjing Li and Xiangyuan Lan and Dongbin Zhao},
    booktitle={The Thirty-ninth Annual Conference on Neural Information Processing Systems},
    year={2026},
    url={https://openreview.net/forum?id=Hs3FrjwyVZ}
}

@inproceedings{tai-etal-2023-exploring,
    title = "Exploring Chain of Thought Style Prompting for Text-to-{SQL}",
    author = "Tai, Chang-Yu  and
      Chen, Ziru  and
      Zhang, Tianshu  and
      Deng, Xiang  and
      Sun, Huan",
    editor = "Bouamor, Houda  and
      Pino, Juan  and
      Bali, Kalika",
    booktitle = "Proceedings of the 2023 Conference on Empirical Methods in Natural Language Processing",
    month = dec,
    year = "2023",
    address = "Singapore",
    publisher = "Association for Computational Linguistics",
    url = "https://aclanthology.org/2023.emnlp-main.327/",
    doi = "10.18653/v1/2023.emnlp-main.327",
    pages = "5376--5393"
}

@inproceedings{lei2025spider,
    title={Spider 2.0: Evaluating Language Models on Real-World Enterprise Text-to-{SQL} Workflows},
    author={Fangyu Lei and Jixuan Chen and Yuxiao Ye and Ruisheng Cao and Dongchan Shin and Hongjin SU and ZHAOQING SUO and Hongcheng Gao and Wenjing Hu and Pengcheng Yin and Victor Zhong and Caiming Xiong and Ruoxi Sun and Qian Liu and Sida Wang and Tao Yu},
    booktitle={The Thirteenth International Conference on Learning Representations},
    year={2025},
    url={https://openreview.net/forum?id=XmProj9cPs}
}

@inproceedings{li2026swesql,
    title={{SWE}-{SQL}: Illuminating {LLM} Pathways to Solve User {SQL} Issues in Real-World Applications},
    author={Jinyang Li and Xiaolong Li and Ge Qu and Per Jacobsson and Bowen Qin and Binyuan Hui and Shuzheng Si and Nan Huo and Xiaohan Xu and Yue Zhang and Ziwei Tang and Yuanshuai Li and Florensia Widjaja and Xintong Zhu and Feige Zhou and Yongfeng Huang and Yannis Papakonstantinou and Fatma Ozcan and Chenhao Ma and Reynold Cheng},
    booktitle={The Thirty-ninth Annual Conference on Neural Information Processing Systems},
    year={2026},
    url={https://openreview.net/forum?id=yRxXTdElLv}
}

@article{guo2024deepseek,
    title={DeepSeek-Coder: when the large language model meets programming--the rise of code intelligence},
    author={Guo, Daya and Zhu, Qihao and Yang, Dejian and Xie, Zhenda and Dong, Kai and Zhang, Wentao and Chen, Guanting and Bi, Xiao and Wu, Yifan and Li, YK and others},
    journal={arXiv preprint arXiv:2401.14196},
    year={2024}
}

@article{qwen25coder,
    title={Qwen2.5-coder technical report},
    author={Hui, Binyuan and Yang, Jian and Cui, Zeyu and Yang, Jiaxi and Liu, Dayiheng and Zhang, Lei and Liu, Tianyu and Zhang, Jiajun and Yu, Bowen and Lu, Keming and others},
    journal={arXiv preprint arXiv:2409.12186},
    year={2024}
}

@inproceedings{spider,
  title={Spider: A large-scale human-labeled dataset for complex and cross-domain semantic parsing and text-to-sql task},
  author={Yu, Tao and Zhang, Rui and Yang, Kai and Yasunaga, Michihiro and Wang, Dongxu and Li, Zifan and Ma, James and Li, Irene and Yao, Qingning and Roman, Shanelle and others},
  booktitle={Proceedings of the 2018 conference on empirical methods in natural language processing},
  pages={3911--3921},
  year={2018}
}

@article{bird,
  title={Can llm already serve as a database interface? a big bench for large-scale database grounded text-to-sqls},
  author={Li, Jinyang and Hui, Binyuan and Qu, Ge and Yang, Jiaxi and Li, Binhua and Li, Bowen and Wang, Bailin and Qin, Bowen and Geng, Ruiying and Huo, Nan and others},
  journal={Advances in Neural Information Processing Systems},
  volume={36},
  pages={42330--42357},
  year={2023}
}

@misc{qwen3coder,
  author       = {{Qwen Team}},
  title        = {Qwen3-Coder: Agentic Coding in the World},
  year         = {2025},
  month        = {July},
  howpublished = {\url{https://qwen.ai/blog?id=qwen3-coder}},
  note         = {Accessed: 2026-05-18}
}

@inproceedings{uncoveringdpo,
  title={Uncovering the impact of chain-of-thought reasoning for direct preference optimization: Lessons from text-to-sql},
  author={Liu, Hanbing and Li, Haoyang and Zhang, Xiaokang and Chen, Ruotong and Xu, Haiyong and Tian, Tian and Qi, Qi and Zhang, Jing},
  booktitle={Proceedings of the 63rd Annual Meeting of the Association for Computational Linguistics (Volume 1: Long Papers)},
  pages={21223--21261},
  year={2025}
}

@article{arctic,
    title={Arctic-text2sql-r1: Simple rewards, strong reasoning in text-to-sql},
    author={Yao, Zhewei and Sun, Guoheng and Borchmann, Lukasz and Nuti, Gaurav and Shen, Zheyu and Deng, Minghang and Zhai, Bohan and Zhang, Hao and Li, Ang and He, Yuxiong},
    journal={arXiv preprint arXiv:2505.20315},
    year={2025}
}

@article{sqlr1,
  title={Sql-r1: Training natural language to sql reasoning model by reinforcement learning},
  author={Ma, Peixian and Zhuang, Xialie and Xu, Chengjin and Jiang, Xuhui and Chen, Ran and Guo, Jian},
  journal={Advances in Neural Information Processing Systems},
  volume={38},
  pages={174505--174537},
  year={2026}
}

@inproceedings{cotnotcot,
  title={To cot or not to cot? chain-of-thought helps mainly on math and symbolic reasoning},
  author={Sprague, Zayne and Yin, Fangcong and Rodriguez, Juan and Jiang, Dongwei and Wadhwa, Manya and Singhal, Prasann and Zhao, Xinyu and Ye, Xi and Mahowald, Kyle and Durrett, Greg},
  booktitle={International Conference on Learning Representations},
  volume={2025},
  pages={94118--94162},
  year={2025}
}

@article{ppo,
  title={Proximal policy optimization algorithms},
  author={Schulman, John and Wolski, Filip and Dhariwal, Prafulla and Radford, Alec and Klimov, Oleg},
  journal={arXiv preprint arXiv:1707.06347},
  year={2017}
}

@article{omnisql,
    author = {Li, Haoyang and Wu, Shang and Zhang, Xiaokang and Huang, Xinmei and Zhang, Jing and Jiang, Fuxin and Wang, Shuai and Zhang, Tieying and Chen, Jianjun and Shi, Rui and Chen, Hong and Li, Cuiping},
    title = {OmniSQL: Synthesizing High-Quality Text-to-SQL Data at Scale},
    year = {2025},
    issue_date = {July 2025},
    publisher = {VLDB Endowment},
    volume = {18},
    number = {11},
    issn = {2150-8097},
    url = {https://doi.org/10.14778/3749646.3749723},
    doi = {10.14778/3749646.3749723},
    journal = {Proc. VLDB Endow.},
    month = jul,
    pages = {4695–4709},
    numpages = {15}
}

@article{codes,
    author = {Li, Haoyang and Zhang, Jing and Liu, Hanbing and Fan, Ju and Zhang, Xiaokang and Zhu, Jun and Wei, Renjie and Pan, Hongyan and Li, Cuiping and Chen, Hong},
    title = {CodeS: Towards Building Open-source Language Models for Text-to-SQL},
    year = {2024},
    issue_date = {June 2024},
    publisher = {Association for Computing Machinery},
    address = {New York, NY, USA},
    volume = {2},
    number = {3},
    url = {https://doi.org/10.1145/3654930},
    doi = {10.1145/3654930},
    journal = {Proc. ACM Manag. Data},
    month = may,
    articleno = {127},
    numpages = {28}
}

@inproceedings{reasoningsql,
    title={Reasoning-{SQL}: Reinforcement Learning with {SQL} Tailored Partial Rewards for Reasoning-Enhanced Text-to-{SQL}},
    author={Mohammadreza Pourreza and Shayan Talaei and Ruoxi Sun and Xingchen Wan and Hailong Li and Azalia Mirhoseini and Amin Saberi and Sercan O Arik},
    booktitle={Second Conference on Language Modeling},
    year={2025},
    url={https://openreview.net/forum?id=HbwkIDWQgN}
}

@article{dpo,
  title={Direct preference optimization: Your language model is secretly a reward model},
  author={Rafailov, Rafael and Sharma, Archit and Mitchell, Eric and Manning, Christopher D and Ermon, Stefano and Finn, Chelsea},
  journal={Advances in neural information processing systems},
  volume={36},
  pages={53728--53741},
  year={2023}
}

@inproceedings{LoRA,
    title={Lo{RA}: Low-Rank Adaptation of Large Language Models},
    author={Edward J Hu and yelong shen and Phillip Wallis and Zeyuan Allen-Zhu and Yuanzhi Li and Shean Wang and Lu Wang and Weizhu Chen},
    booktitle={International Conference on Learning Representations},
    year={2022},
    url={https://openreview.net/forum?id=nZeVKeeFYf9}
}

@inproceedings{zheng2024llamafactory,
  title={LlamaFactory: Unified Efficient Fine-Tuning of 100+ Language Models},
  author={Yaowei Zheng and Richong Zhang and Junhao Zhang and Yanhan Ye and Zheyan Luo and Zhangchi Feng and Yongqiang Ma},
  booktitle={Proceedings of the 62nd Annual Meeting of the Association for Computational Linguistics (Volume 3: System Demonstrations)},
  address={Bangkok, Thailand},
  publisher={Association for Computational Linguistics},
  year={2024},
  url={http://arxiv.org/abs/2403.13372}
}

\appendix
\section{Implementation Details}\label{app:impl}

\paragraph{SFT Training.}
Both SFT and DPO are implemented using the LLaMA-Factory framework~\citep{zheng2024llamafactory}.
We fine-tune Qwen3-Coder-30B-A3B-Instruct using LoRA~\citep{LoRA} with rank $r{=}16$, $\alpha{=}32$, and dropout $0.05$, applied to all attention and MLP projection layers (\texttt{q,k,v,o,gate,up,down\_proj}).
We use the AdamW optimizer with a cosine learning rate schedule, a peak learning rate of $1 \times 10^{-4}$, and a warmup ratio of $0.1$.
Training runs for $3$ epochs with an effective batch size of $64$ and a maximum sequence length of $4096$.
SFT is conducted for approximately 20 minutes with $4\times$ NVIDIA H100 (80GB) GPUs using DeepSpeed ZeRO-2 and bfloat16 precision.

\paragraph{DPO Training.}
For hyperparameters on DPO training, we use lora rank $r{=}16$, $\alpha{=}32$, dropout $0.05$, cosine learning rate schedule with a peak learning rate of $1 \times 10^{-4}$ and warmup ratio $0.1$, effective batch size $64$, and maximum sequence length $4096$ and $\beta$ $0.3$.
DPO is conducted on $4\times$ NVIDIA H100 (80GB) GPUs with DeepSpeed ZeRO-3 and bfloat16 precision. Training hours consumed are approximately 11 hours, 9 hours, and 4 hours for AutoThinkSQL, CoT-only, no-CoT-only model, respectively.

\paragraph{Inference.}
All inference is performed on $4\times$ NVIDIA L40S GPUs using vLLM for efficient batched generation.
For Maj@8, we sample $8$ independent outputs per query with temperature $1.0$ and top-$p$ $1.0$, and select the final answer by majority vote over execution outcomes.
Evaluation follows the codebase at \url{https://github.com/QwenLM/Qwen3-Coder/tree/main/qwencoder-eval}, applying the mode-specific format instruction appropriate for each evaluated model.

\section{Dataset Preprocessing}\label{app:preprocessing}
\paragraph{Data Schema Pre-Processing.}
In this work, we utilize the preprocessed dataset provided by ~\cite{uncoveringdpo}, which follows the preprocessing methodology introduced in CodeS~\citep{codes}. 
Specifically, a schema item classifier is first employed to retrieve the tables and columns relevant to a given question. 
Subsequently, values related to the questions are extracted using a coarse-to-fine approach. 
Once all the necessary data components are successfully extracted, they are integrated to construct the final database prompt.

\paragraph{Dataset Details.}
For SFT dataset construction, we collect $16$ independent rollouts under both $p_{\text{NC}}$ (Figure~\ref{fig:format_nocot}) and $p_{\text{C}}$ (Figure~\ref{fig:format_cot}) for each BIRD training instance, as described in Section~\ref{ssec22:sft}.
Responses generated under $p_{\text{NC}}$ are wrapped in \texttt{[SQL]\,\ldots\,[/SQL]} tags before being included in the final training data.
Because AutoThinkSQL leverages rollouts from both modes, it can retain more training instances than CoT-only or No-CoT-only baselines, which rely on a single mode and must discard any instance that yields no correct response under that mode.
The resulting SFT data statistics are summarized in Table~\ref{tab:sft_data_stats}.

For DPO dataset construction, we similarly collect $16$ new rollouts per instance under $p_{\text{NC}}$ (Figure~\ref{fig:format_nocot}) and $p_{\text{C}}$ (Figure~\ref{fig:format_cot}) using the SFT-trained model, while CoT-only and No-CoT-only baselines collect rollouts under only their respective format instruction.
Auto-thinking preference pairs are constructed as explained in Section~\ref{ssec23:dpo}; instances with all-correct or all-incorrect responses are discarded.
The resulting DPO data sizes are shown in Table~\ref{tab:dpo_data_stats}, with AutoThinkSQL retaining more pairs by utilizing rollouts from both $p_{\text{NC}}$ and $p_{\text{C}}$.

\begin{table}[H]
    \centering
    \small
    \caption{Number of SFT training dataset including the breakdown of CoT and No-CoT pairs.}
    \label{tab:sft_data_stats}
    \resizebox{0.75\columnwidth}{!}{%
    \begin{tabular}{l r r r}
        \toprule
        & & \multicolumn{2}{c}{\textbf{Breakdown}} \\
        \cmidrule(lr){3-4}
        \textbf{Type} & \textbf{Total \#} & \textbf{CoT \#} & \textbf{No-CoT \#} \\
        \midrule
        AutoThinkSQL  & 5981 & 4625 & 1356 \\
        CoT-Only      & 5873 & 5873 & 0 \\
        No-CoT-Only   & 5352 & 0    & 5352 \\
        \bottomrule
    \end{tabular}%
    }
\end{table}
\begin{table}[H]
    \centering
    \small
    \caption{Number of valid DPO training pairs generated by respective SFT models.}
    \label{tab:dpo_data_stats}
    \resizebox{0.6\columnwidth}{!}{%
    \begin{tabular}{l r}
        \toprule
        \textbf{SFT Model type} & \textbf{Total Pair \#} \\
        \midrule
        AutoThinkSQL      & 6064 \\
        CoT-Only      & 4896 \\
        No-CoT-Only   & 2425 \\
        \bottomrule
    \end{tabular}%
    }
\end{table}
\begin{table}[H]
    \centering
    \footnotesize
    \setlength{\tabcolsep}{2pt}
    \caption{Licenses and sources of scientific artifacts.}
    \label{tab:licenses}
    \resizebox{\columnwidth}{!}{%
    \begin{tabular}{llll}
        \toprule
        \textbf{Artifact} & \textbf{Type} & \textbf{License} & \textbf{Source} \\
        \midrule
        Qwen3-Coder-30B-A3B~\citeyearpar{qwen3coder} & Model & Apache 2.0 & \href{https://huggingface.co/Qwen/Qwen3-Coder-30B-A3B-Instruct}{link} \\
        Spider~\citeyearpar{spider} & Dataset & CC BY-SA 4.0 & \href{https://yale-lily.github.io/spider}{link} \\
        BIRD~\citeyearpar{bird} & Dataset & CC BY-SA 4.0 & \href{https://bird-bench.github.io/}{link} \\
        LLaMA-Factory~\citeyearpar{zheng2024llamafactory} & Framework & Apache 2.0 & \href{https://github.com/hiyouga/LLaMA-Factory}{link} \\
        DeepSpeed & Framework & Apache 2.0 & \href{https://github.com/microsoft/DeepSpeed}{link} \\
        vLLM & Framework & Apache 2.0 & \href{https://github.com/vllm-project/vllm}{link} \\
        \bottomrule
    \end{tabular}
    }
\end{table}

\section{Prompts}\label{sec:appendix_prompts}
Figure~\ref{fig:shared_template} shows the shared prompt template used for all SFT and DPO training instances.
The template combines a task overview, the preprocessed database schema, matched contents identified by schema linking, optional external knowledge, and the natural language question, followed by a mode-specific format instruction.
Figure~\ref{fig:format_ours} presents the auto-thinking format instruction used by AutoThinkSQL, which instructs the model to first judge query complexity and then either emit the SQL directly or produce reasoning before the SQL.
For comparison, Figure~\ref{fig:format_cot} and Figure~\ref{fig:format_nocot} show the CoT-only and No-CoT-only format instructions used by the corresponding baselines.

\begin{figure}
\centering
\newcommand{\promptvar}[1]{\textcolor{black!50}{\textit{<#1>}}}
\newcommand{\promptvarbracket}[1]{\textbf{[#1]}}
\begin{tcolorbox}[
    enhanced jigsaw,
    colback=blue!2!white,
    colframe=blue!40!black,
    boxrule=0.6pt,
    titlerule=0.4pt,
    toptitle=3pt, bottomtitle=3pt,
    fonttitle=\bfseries,
    colbacktitle=blue!8!white,
    coltitle=black,
    title=Prompt Template,
    arc=2pt,
    left=5pt, right=5pt, top=5pt, bottom=5pt,
    width=\columnwidth
]
\footnotesize
\begin{Verbatim}[breaklines=true, breaksymbol={}, commandchars=\\\|\~]
Task Overview:
You are a helpful SQL expert assistant. Below, you are provided with a database schema and a natural language question. Your task is to understand the schema and generate a valid SQL query to answer the question.

Database Engine:
SQLite

Database Schema:
\promptvar|SCHEMA_STR~

Matched Contents:
\promptvar|MATCHED_STR~

External Knowledge:
\promptvar|EVIDENCE~

Question:
\promptvar|QUESTION~

\promptvarbracket|FORMAT INSTRUCTION~
\end{Verbatim}
\end{tcolorbox}
\caption{Shared prompt template for SFT and DPO training instances.}
\label{fig:shared_template}
\end{figure}

\begin{figure}[H]
\centering
\newcommand{\promptvar}[1]{\textcolor{black!50}{\textit{<#1>}}}
\begin{tcolorbox}[
    enhanced jigsaw,
    colback=blue!2!white,
    colframe=blue!40!black,
    boxrule=0.6pt,
    titlerule=0.4pt,
    toptitle=3pt, bottomtitle=3pt,
    fonttitle=\bfseries,
    colbacktitle=blue!8!white,
    coltitle=black,
    title=Format Instruction (CoT),
    arc=2pt,
    left=5pt, right=5pt, top=5pt, bottom=5pt,
    width=\columnwidth
]
\footnotesize
\begin{Verbatim}[breaklines=true, breaksymbol={}, commandchars=\\\|\~]
Please think step-by-step and output the final SQL query between [SQL] and [/SQL].
\end{Verbatim}
\end{tcolorbox}
\caption{CoT-only format instruction.}
\label{fig:format_cot}
\end{figure}

\section{Licenses}\label{app:licenses}
We list the licenses and sources of all scientific artifacts used in this work in Table~\ref{tab:licenses}. 
All artifacts are publicly released under permissive licenses that allow research use, and our usage is consistent with their intended purposes.

\section{Use of AI Assistants}\label{ai-assistants}
We used AI assistants (ChatGPT, Claude, and Gemini) for both writing assistance (grammar, phrasing, and translation) and coding support (debugging and boilerplate code). 
All research ideas, experimental design, and analyses were conducted by the authors, and all AI-assisted content was reviewed and verified.

\begin{figure}[H]
\centering
\newcommand{\promptvar}[1]{\textcolor{black!50}{\textit{<#1>}}}
\begin{tcolorbox}[
    enhanced jigsaw,
    colback=blue!2!white,
    colframe=blue!40!black,
    boxrule=0.6pt,
    titlerule=0.4pt,
    toptitle=3pt, bottomtitle=3pt,
    fonttitle=\bfseries,
    colbacktitle=blue!8!white,
    coltitle=black,
    title=Format Instruction (No-CoT),
    arc=2pt,
    left=5pt, right=5pt, top=5pt, bottom=5pt,
    width=\columnwidth
]
\footnotesize
\begin{Verbatim}[breaklines=true, breaksymbol={}, commandchars=\\\|\~]
Please output only the final SQL query, starts with keyword `SELECT`.
\end{Verbatim}
\end{tcolorbox}
\caption{No-CoT-only format instruction.}
\label{fig:format_nocot}
\end{figure}

\begin{figure}[H]
\centering
\newcommand{\promptvar}[1]{\textcolor{black!50}{\textit{<#1>}}}
\begin{tcolorbox}[
    enhanced jigsaw,
    colback=blue!2!white,
    colframe=blue!40!black,
    boxrule=0.6pt,
    titlerule=0.4pt,
    toptitle=3pt, bottomtitle=3pt,
    fonttitle=\bfseries,
    colbacktitle=blue!8!white,
    coltitle=black,
    title=Format Instruction (AutoThinkSQL),
    arc=2pt,
    left=5pt, right=5pt, top=5pt, bottom=5pt,
    width=\columnwidth
]
\footnotesize
\begin{Verbatim}[breaklines=true, breaksymbol={}, commandchars=\\\|\~]
Format Instructions:
Analyze the question, database schema, and external knowledge to determine how to write the SQL query.
First, evaluate the complexity of the problem.
If it is an easy problem: Do not provide any reasoning or explanation. Directly output the final SQL query enclosed within [SQL] and [/SQL] tags.
If it is a difficult problem requiring thought: Write down your detailed reasoning and analysis process first as plain text. After you have thoroughly thought through the problem, provide the final SQL query enclosed within [SQL] and [/SQL] tags at the very end.

Output format:
(For easy problems)
[SQL]
Your SQL here
[/SQL]

(For difficult problems)
Your reasoning here
[SQL]
Your SQL here
[/SQL]
\end{Verbatim}
\end{tcolorbox}
\caption{Auto-thinking format instruction.}
\label{fig:format_ours}
\end{figure}

\end{document}